%% file: main.tex
\documentclass[10pt,twocolumn,letterpaper]{article}

\usepackage{iccv}              

\input{preamble}

\definecolor{iccvblue}{rgb}{0.21,0.49,0.74}
\definecolor{Gray}{gray}{0.85}
\usepackage[pagebackref,breaklinks,colorlinks,allcolors=iccvblue]{hyperref}

\def\paperID{7060} 
\def\confName{ICCV}
\def\confYear{2025}

\title{EVT: Efficient View Transformation for Multi-Modal 3D Object Detection}

\author{Yongjin Lee\textsuperscript{1}, Hyeon-Mun Jeong\textsuperscript{1}, Yurim Jeon\textsuperscript{2}, Sanghyun Kim\textsuperscript{1,2}\thanks{Corresponding Author.} \\
\normalsize$^1$ThorDrive Co., Ltd, South Korea \quad \normalsize$^2$Seoul National University, South Korea \\
{\tt\small \{dydwls462, hyeonmun123\}@gmail.com}\quad{\tt\small \{fabioisyo01, shyun613\}@snu.ac.kr}
}

\begin{document}
\maketitle

\input{sec/0_abstract}
\vspace{-2pt}
\input{sec/1_introduction}
\input{sec/2_related_work}
\input{sec/3_methodology}
\input{sec/4_experiments}
\input{sec/5_conclusions}

{
    \small
    \bibliographystyle{ieeenat_fullname}
    \bibliography{main}
}

\end{document}

%% file: preamble.tex
\usepackage{color, colortbl}
\usepackage{graphicx}
\usepackage{subcaption}
\usepackage{multirow}


%% file: sec/0_abstract.tex
\begin{abstract}

Multi-modal sensor fusion in Bird’s Eye View (BEV) representation has become the leading approach for 3D object detection. However, existing methods often rely on depth estimators or transformer encoders to transform image features into BEV space, which reduces robustness or introduces significant computational overhead. Moreover, the insufficient geometric guidance in view transformation results in ray-directional misalignments, limiting the effectiveness of BEV representations.
To address these challenges, we propose Efficient View Transformation (EVT), a novel 3D object detection framework that constructs a well-structured BEV representation, improving both accuracy and efficiency. Our approach focuses on two key aspects. First, Adaptive Sampling and Adaptive Projection (ASAP), which utilizes LiDAR guidance to generate 3D sampling points and adaptive kernels, enables more effective transformation of image features into BEV space and a refined BEV representation. Second, an improved query-based detection framework, incorporating group-wise mixed query selection and geometry-aware cross-attention, effectively captures both the common properties and the geometric structure of objects in the transformer decoder.
On the nuScenes test set, EVT achieves state-of-the-art performance of 75.3\% NDS with real-time inference speed.

\end{abstract}

%% file: sec/1_introduction.tex
\section{Introduction}
\label{sec:intro}
LiDAR-camera fusion is essential for 3D object detection, as it leverages the complementary strengths \cite{Li2022uvtr, chen2023futr3d, yin2021multimodal_mvp, vora2020pointpainting, liu2023bevfusion, liang2022bevfusion, Bai2022TransFusionRL, chen2023focalformer3d, yang2022deepinteraction, xie2023sparsefusion, Yan2023cmt, cai2023bevfusion4d}. LiDAR provides precise geometric information for accurate object localization, while cameras capture rich semantic details such as color and texture. However, their integration remains challenging due to the differences in sensing modalities.

\begin{figure}[t]
    \centering
    \includegraphics[width=\linewidth]{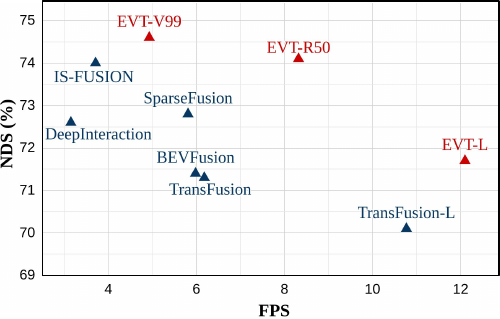}
    \vspace{-20pt}
    \caption{Performance comparison of EVT and other methods on the nuScenes validation set. The FPS of all methods is measured in FP32 on a single Tesla A100 GPU using the official implementations, excluding voxelization time.
    }
    \vspace{-15pt}
\label{fig:intro}
\end{figure}

Recently, the dominant multi-modal fusion methods are categorized as implicit or explicit fusion. Implicit fusion employs cross-attention within transformers, where object queries iteratively interact with independently processed sensor features \cite{chen2023futr3d, Bai2022TransFusionRL, yang2022deepinteraction, yang2024deepinteraction++, Yan2023cmt}. Implicit fusion offers flexibility and simplicity by removing explicit feature alignment across sensor modalities but incurs high computational costs and struggles to extract complementary features. Explicit fusion, on the other hand, directly aligns and integrates sensor data in BEV space using view transformation (VT), which enhances complementary feature fusion while reducing computational costs. Although its performance heavily depends on the accuracy of VT, conventional methods have inherent limitations. Depth-based VT methods \cite{philion2020lift_splat_shot, li2023bevdepth, liu2023bevfusion, liang2022bevfusion, zhao2024simplebev, li2024gafusion} lift image features into BEV using pixel-wise depth estimation, but their sensitivity to depth errors compromises robustness. In contrast, query-based VT methods \cite{li2022bevformer, Bai2022TransFusionRL, chen2022autoalign, chen2022autoalignv2, cai2023bevfusion4d, hu2023fusionformer} refine BEV queries via attention mechanisms, incurring high computational costs. Furthermore, both methods lack geometric guidance, resulting in ray-directional misalignment. This misalignment reduces spatial accuracy, leading to unintended information capture along the ray direction, ultimately degrading BEV representation accuracy.

To address the limitations of explicit fusion methods and improve detection performance, we propose Efficient View Transformation (EVT), a novel 3D object detection framework aimed at enhancing both accuracy and efficiency by constructing a well-structured BEV representation. EVT introduces two key innovations: (1) Adaptive Sampling and Adaptive Projection (ASAP), a novel VT method which leverages LiDAR guidance, and (2) an improved query-based object detection framework incorporating group-wise mixed query selection and geometry-aware cross-attention, enabling effective multi-modal BEV feature decoding for accurate 3D object detection.

ASAP consists of two key modules: Adaptive Sampling (AS) and Adaptive Projection (AP). AS generates 3D sampling points from LiDAR features to effectively represent image features in BEV space while focusing more on high-relevance areas in the image. AP refines BEV representations using adaptive kernels generated from LiDAR features to enhance structurally meaningful 3D information. As a result, ASAP improves BEV feature representation and eliminates ray-directional misalignment. Unlike existing methods, it does not rely on depth estimation or attention mechanism, ensuring both efficiency and robustness.

Additionally, we improve the query-based object detection framework by introducing group-wise mixed query selection and geometry-aware cross-attention. The mixed query selection generates object queries using group-wise learnable parameters and heatmaps, allowing them to capture the common properties of each group and initialize at high-confidence positions. Then, the geometry-aware cross-attention refines object queries by integrating corner-aware sampling for precise feature selection and position-aware feature mixing for spatially aware feature decoding. These enhancements improve more robust and accurate detection while maintaining computational efficiency.

We evaluate EVT on the nuScenes dataset in terms of accuracy and efficiency. As shown in \cref{fig:intro}, EVT achieves 74.1\% NDS and 8.3 FPS with ResNet-50 \cite{resnet}, 74.6\% NDS and 4.9 FPS with V2-99 \cite{Lee2019VoVNetV2}, and 71.7\% NDS and 12.1 FPS with the LiDAR-only model EVT-L on the nuScenes validation set, outperforming other methods in both accuracy and inference speed. On the nuScenes test set, EVT achieves 75.3\% NDS and 72.5\% mAP using single-frame raw data, without model ensemble or test-time augmentation, surpassing previous state-of-the-art methods.

In summary, the main contributions are as follows:
\begin{itemize}
    \item We introduce EVT, a novel 3D object detection framework that improves both accuracy and efficiency through ASAP and an improved query-based framework.
    \item ASAP, our view transformation method, utilizes LiDAR guidance to generate BEV feature maps while improving efficiency by eliminating the need for depth estimators and transformer encoders.
    \item Our improved query-based detection framework further enhances detection accuracy and robustness.
    \item EVT achieves state-of-the-art performance of 75.3\% NDS and 72.6\% mAP on the nuScenes test set.
\end{itemize}

%% file: sec/2_related_work.tex
\section{Related Work}
\label{sec:related_work}

\begin{figure*}[t]
    \centering
    \includegraphics[width=\linewidth]{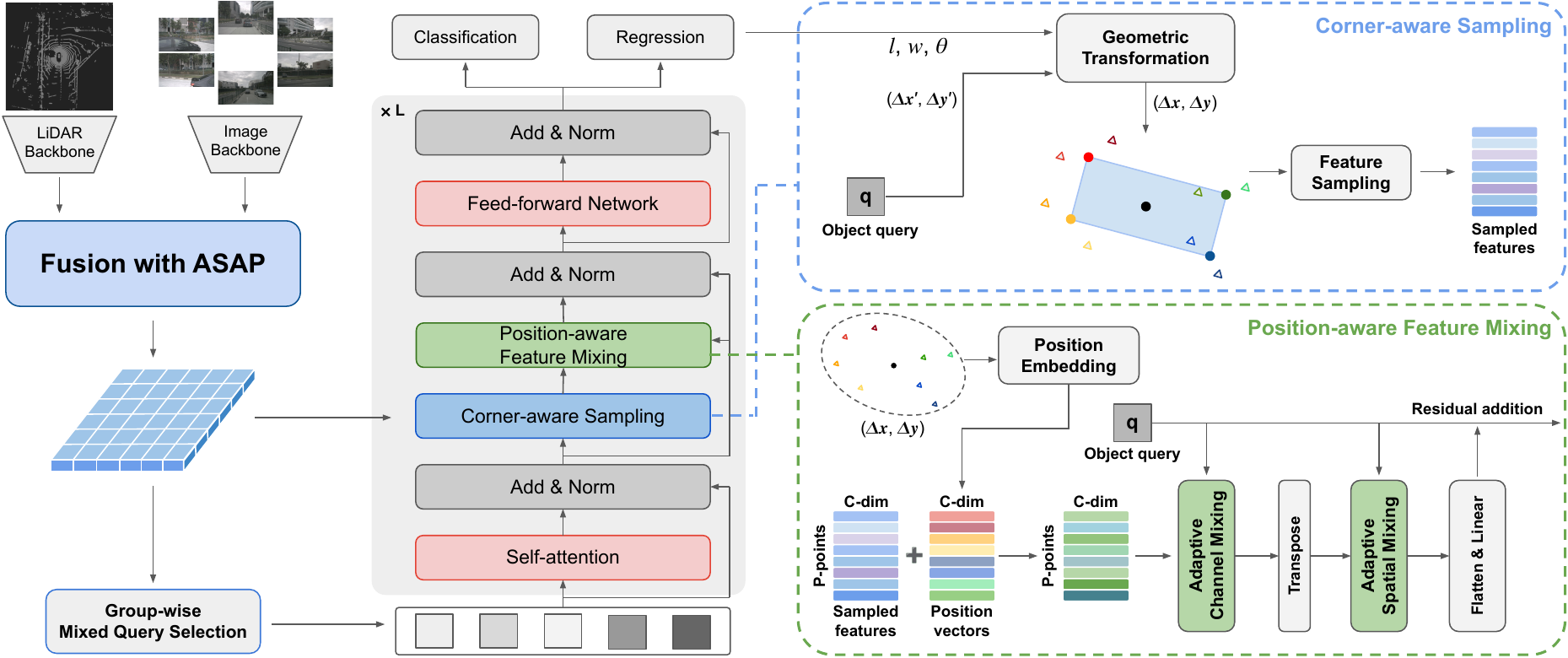}
    \vspace{-15pt}
    \caption{Overall architecture of EVT. Each backbone extracts either image features or LiDAR features. The proposed ASAP module fuses these two features in BEV space. The group-wise mixed query selection stage generates queries from the fused feature map. In the transformer decoder, corner-aware sampling leverages the geometric properties of the object queries to sample multi-modal features, and position-aware feature mixing decodes the sampled features to update the queries. These queries then predict the 3D bounding boxes.
    }
    \label{fig:architecture}
\end{figure*}

\subsection{Query-based Object Detection Framework}
DETR~\cite{erabati2023li3detr} introduces transformers into 2D object detection, eliminating hand-designed components like non-maximum suppression, but suffering from slow convergence. To address this, Deformable DETR~\cite{Zhu2020DeformableDetr} proposes deformable cross-attention to accelerate training. DAB-DETR~\cite{liu2022dab_detr} enhances query representation by modeling queries as anchor boxes, while DN-DETR~\cite{li2022dn_detr} stabilizes training with query denoising and auxiliary supervision.

DETR-like approaches have been extended to 3D object detection \cite{Wang2021detr3d, liu2022petr, liu2023petrv2, luo2022detr4d, wang2023stream_petr, jiang2024far3d, liu2023sparsebev, chen2023futr3d, Li2022uvtr, yang2022deepinteraction, Yan2023cmt}. PETR~\cite{liu2022petr} defines object query features and positions using learnable parameters, leveraging position embeddings within multi-head attention for feature refinement. DETR3D~\cite{Wang2021detr3d} and BEVFormer~\cite{li2022bevformer} project BEV queries onto image planes, refining them with bilinear interpolation or deformable cross-attention. CenterFormer~\cite{zhou2022centerformer} samples features from high-scoring heatmap keypoints for query initialization, while TransFusion~\cite{Bai2022TransFusionRL} enhances sampled features with category embeddings.

While extensive research has explored query initialization, DINO~\cite{zhang2022dino} notes that direct feature sampling in methods \cite{zhou2022centerformer, Bai2022TransFusionRL, yin2024fusion, zhang2024sparselif, wang2024mv2dfusion} limits performance due to inaccurate initial feature representations. Furthermore, despite advancements in attention-based query refinement, these approaches insufficiently leverage the geometric structure of object queries, limiting 3D spatial understanding.

\subsection{Implicit Multi-modal Fusion in Transformer}
Implicit multi-modal fusion integrates each sensor’s data through cross-attention with object queries in the transformer decoder, without relying on BEV representation for feature alignment. FUTR3D~\cite{chen2023futr3d} introduces a modality-agnostic feature sampler that aggregates features from different sensors using deformable cross-attention \cite{Zhu2020DeformableDetr}. DeepInteraction~\cite{yang2022deepinteraction} and DeepInteraction++~\cite{yang2024deepinteraction++} preserve modality-specific information by using a modality interaction strategy. Meanwhile, TransFusion~\cite{Bai2022TransFusionRL} updates object queries in a sequential manner by applying cross-attention separately to LiDAR and camera features. And CMT~\cite{Yan2023cmt} constructs input tokens by adding 3D position embeddings to each sensor's data before concatenating them.

These implicit fusion methods rely on cross-attention between object queries and sensor features within the transformer decoder, providing a flexible and generalizable framework for multi-modal fusion. However, because it depends entirely on query-driven attention mechanisms, it struggles to effectively extract complementary features across different sensors. Furthermore, the iterative execution of cross-attention over the entire sensor data incurs high computational overhead and memory inefficiency.

\subsection{Explicit Multi-modal Fusion in BEV Space}
Explicit multi-modal fusion focuses on view transformation of 2D image features into BEV space \cite{Li2022uvtr, liu2023bevfusion, liang2022bevfusion, Bai2022TransFusionRL, cai2023bevfusion4d, reading2021categorical, philion2020lift_splat_shot, huang2021bevdet, huang2022bevdet4d, li2023bevdepth, li2024fast_bev, li2022bevformer, yang2023bevformer_v2}. For view transformation, depth-based methods predict pixel-wise depth distributions to lift multi-view image features into BEV space \cite{philion2020lift_splat_shot, huang2021bevdet, huang2022bevdet4d, li2023bevdepth, Li2022uvtr, liu2023bevfusion, liang2022bevfusion}. While these methods effectively incorporate spatial priors through depth estimation, their heavy reliance on the depth estimators significantly limits their overall robustness.

In contrast, query-based view transformation methods \cite{li2022bevformer, yang2023bevformer_v2, cai2023bevfusion4d, hu2023fusionformer} project predefined 3D points onto image planes and extract features using deformable cross-attention \cite{Zhu2020DeformableDetr}. This approach eliminates the need for depth estimation; however, predefined positions of sampling points fail to align accurately with regions where objects are located,
leading to suboptimal feature representation. Furthermore, it suffers from high computational overhead due to the extensive use of multi-layer transformers and ray-directional misalignment due to insufficient geometric guidance.

Aforementioned challenges cause the inherent difficulty in establishing precise correspondences between 2D and 3D spaces, which is crucial for effective BEV representation. Existing approaches often suffer from feature misalignment due to depth estimation errors and also struggle with maintaining spatial consistency in transformer-based methods, while their high computational cost significantly limits real-time deployment. Therefore, a more efficient and geometrically grounded approach is essential for achieving accurate and robust multi-modal BEV representations.

%% file: sec/3_methodology.tex
\section{Methodology}
The overall pipeline of EVT is illustrated in \cref{fig:architecture}. First, $N_{\text{s}}$-scale perspective-view image features $\{\mathbf{PV}_j\}_{j=1}^{N_{\text{s}}}$ and BEV LiDAR features $\mathbf{BEV}_{\text{lidar}} \in \mathbb{R}^{C \times H \times W}$ are extracted from separate backbone networks, where $C$ is the feature dimension and $H \times W$ is the size of the BEV feature map.

To fuse 2D image and LiDAR features in BEV space, the Adaptive Sampling and Adaptive Projection (ASAP) module transforms 2D image features into BEV space and then fuses them with $\mathbf{BEV}_{\text{lidar}}$ (\cref{method:asap}).
For query-based 3D detection with multi-modal BEV features, group-wise mixed query selection initializes object queries based on heatmap-guided locations (\cref{method:query_init}), and geometry-aware cross-attention refines them in the transformer decoder to predict 3D bounding boxes (\cref{method:query_update}).

\subsection{Adaptive Sampling and Adaptive Projection}
\label{method:asap}
\begin{figure}[t]
    \centering
    \includegraphics[width=\linewidth]{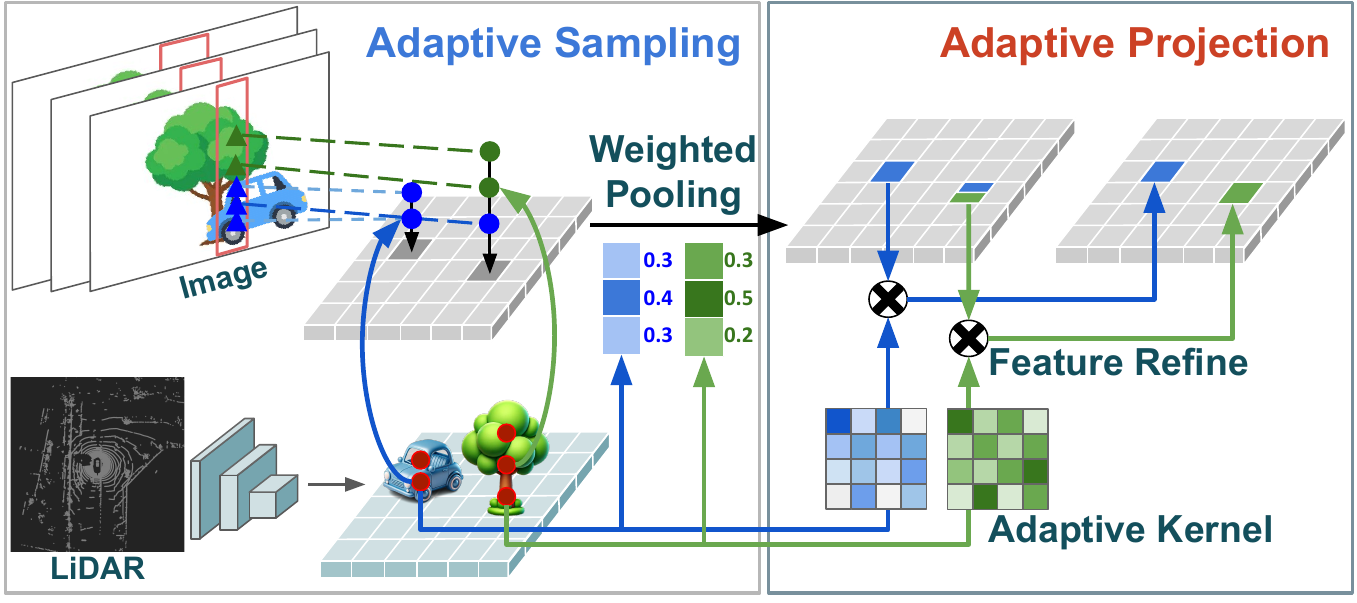}
    \vspace{-17pt}
    \caption{Overview of AS and AP. In AS, the LiDAR BEV feature map generates 3D sampling points and their corresponding weights. The sampling points are projected onto the image plane, where features are sampled and combined by weighted pooling in each BEV grid cell. In AP, the image BEV feature map produced by AS is refined channel-wise using the adaptive kernels generated from the LiDAR features.
    }
    \vspace{-3pt}
\label{fig:asap_archi}
\end{figure}

The proposed ASAP module consists of two stages: Adaptive Sampling (AS) and Adaptive Projection (AP). In the first stage, AS selectively extracts and aggregates multi-scale perspective-view image features $\{\mathbf{PV}_j\}_{j=1}^{N_{\text{s}}}$ into an initial BEV representation $\mathbf{BEV}_{\text{as}}$. In the second stage, AP refines $\mathbf{BEV}_{\text{as}}$ to obtain the final image BEV feature map $\mathbf{BEV}_{\text{camera}}$. The structure of this module is shown in \cref{fig:asap_archi}.

\vspace{-10pt}
\paragraph{Adaptive Sampling}
To transform multi-scale image features into BEV space, the Adaptive Sampling (AS) module predicts optimal sampling heights for each BEV grid cell using LiDAR features, allowing effective feature extraction from high-relevance areas in the image (see \cref{fig:asap_proj}).

First, multiple height values for each grid cell are generated from LiDAR features:
\begin{equation}
    \{Z_i\}_{i=1}^{N_{\text{h}}} = {\rm Conv}(\mathbf{BEV}_{\text{lidar}})(u, v),
\end{equation}
where $(u, v)$ denotes the coordinates of a grid cell in BEV space, and $N_{\text{h}}$ denotes the number of generated heights. Consequently, the 3D sampling points $P = \{(X, Y, Z_i)\}_{i=1}^{N_\text{h}}$ are defined based on real-world coordinates $(X, Y)$ corresponding to the grid cell at $(u, v)$ and the set of heights $\{Z_i\}$.

Next, the generated points $P$ are projected onto the image plane. Each projected point samples the multi-scale features $\{\mathbf{PV}_j\}_{j=1}^{N_{\text{s}}}$ across $N_{\text{s}}$ different image scales with downsampled strides $\{S_j\}_{j=1}^{N_{\text{s}}}$. Consequently, $N_{\text{h}} \times N_{\text{s}}$ projected points $(x^{j}_{i}, y^{j}_{i})$ and sampled features $f^{j}_{i}$ are obtained via projection:
\begin{gather}
    (x^{j}_{i}, y^{j}_{i}) = \frac{{\rm Proj} (X, Y, Z_i)}{S_{j}} \\
    f^{j}_{i} = {\rm B}(\mathbf{PV}_{j}, (x^{j}_{i}, y^{j}_{i})) \in \mathbb{R}^{C},
\end{gather}
where ${\rm Proj}(\cdot)$ denotes the projection of 3D points onto the image plane, and ${\rm B}(\cdot)$ denotes bilinear interpolation.

To aggregate the $N_{\text{s}} \times N_{\text{h}}$ sampled features $\{f^{j}_{i}\}$, adaptive sampling weights $W_{\text{as}} \in \mathbb{R}^{N_{\text{s}} \times N_{\text{h}}}$ are derived from $\mathbf{BEV}_{\text{lidar}}$. $W_{\text{as}}$ determines the importance of the heights and image feature scales for each grid cell. The multi-scale image features in BEV space are obtained as follows:
\begin{gather}
    W_{\text{as}} = \sigma({\rm Conv}(\mathbf{BEV}_{\text{lidar}})(u, v)) \\
    \mathbf{BEV}_{\text{as}}(u, v) = \sum_{j=1}^{N_{\text{s}}} \sum_{i=1}^{N_{\text{h}}} W_{\text{as}}(j, i) \cdot f^{j}_{i} \in \mathbb{R}^{C},
\end{gather}
where $\sigma(\cdot)$ denotes the softmax function applied to all $N_\text{s} \times N_\text{h}$ elements. $\mathbf{BEV}_{\text{as}}$ denotes the image BEV feature map.

\paragraph{Adaptive Projection}
The Adaptive Projection (AP) module refines the BEV feature map $\mathbf{BEV}_{\text{as}}$, produced by the AS module, by applying adaptive kernels to each grid cell. The overall process is represented by the following equations:
\begin{gather}
    K_{\text{ap}} = {\rm Conv}(\mathbf{BEV}_{\text{lidar}})(u, v) \in \mathbb{R}^{C \times C} \\
    \mathbf{BEV}_{\text{camera}}(u, v) = \mathbf{BEV}_{\text{as}}(u, v) \times K_{\text{ap}} \in \mathbb{R}^{C}.
\end{gather}

First, an adaptive kernel $K_{\text{ap}} \in \mathbb{R}^{C \times C}$ is derived from $\mathbf{BEV}_{\text{lidar}}$ for each BEV grid cell. Then, $\mathbf{BEV}_{\text{as}}$ is refined by applying a channel-wise linear projection using $K_{\text{ap}}$ to obtain the image BEV feature map $\mathbf{BEV}_{\text{camera}}$.

Unlike static transformations, our LiDAR-guided feature refinement leverages spatial information to effectively mitigate ray-directional misalignment, which primarily results from occlusions and empty 3D spaces (see \cref{fig:asap_bev}).

\paragraph{Multi-modal Fusion in BEV Space}
The multi-modal BEV feature map $\mathbf{BEV}_{\text{fuse}} \in \mathbb{R}^{C \times H \times W}$ is obtained by concatenating the image and LiDAR BEV feature maps ($\mathbf{BEV}_{\text{camera}}$ and $\mathbf{BEV}_{\text{lidar}}$) along the channel dimension, followed by a convolution operation:
\begin{equation}
\mathbf{BEV}_{\text{fuse}} = {\rm Conv}({\rm Concat}(\mathbf{BEV}_{\text{camera}}, \mathbf{BEV}_{\text{lidar}})),
\end{equation}
where ${\rm Concat}(\cdot)$ denotes channel-wise concatenation.

\subsection{Group-wise Mixed Query Selection}
\label{method:query_init}
Our proposed group-wise mixed query selection generates object queries for transformer-based detection frameworks. First, group-wise heatmaps are predicted from the multi-modal BEV feature map $\mathbf{BEV}_{\text{fuse}}$ (from \cref{method:asap}), where each group consists of similarly sized object classes. The predicted heatmaps have scores ranging from 0 to 1, representing the likelihood that each BEV pixel corresponds to the center of an object. The heatmap head is supervised by 2D Gaussian distributions centered at each object's location. Next, the top-k keypoints are selected from each heatmap group, and their positions are used as reference points for queries in BEV space.

Inspired by DINO~\cite{zhang2022dino}, we initialize query features solely with group-wise learnable parameters, unlike query positions. DINO defines query features as instance-wise learnable parameters without using any categorical priors. In contrast, our approach allows all queries within the same group to share these parameters, effectively capturing the common properties of their group. Experimentally, the group-wise shared initial embeddings outperform instance-wise embeddings for object query representation. Moreover, our approach outperforms traditional approaches that either define both query features and positions as learnable parameters \cite{liu2022petr, liu2023petrv2, liu2023sparsebev, Yan2023cmt, hu2023fusionformer} or obtain both from heatmap keypoints \cite{zhou2022centerformer, Bai2022TransFusionRL, yin2024fusion, zhang2024sparselif, wang2024mv2dfusion}. Further details and analysis of our method are provided in \cref{sec:ablation}.

\subsection{Geometry-aware Cross-Attention}
\label{method:query_update}
To enhance query representations in transformer decoders, we refine deformable cross-attention \cite{Zhu2020DeformableDetr} with corner-aware sampling for improved feature sampling and position-aware feature mixing for better feature aggregation.

\vspace{-10pt}
\paragraph{Corner-aware Sampling}
The conventional deformable cross-attention samples features around the centers of object queries. However, this approach struggles with objects of varying sizes and fails to capture fine-grained boundary details and spatial extent effectively. To address this limitation, corner-aware sampling explicitly incorporates object geometry, ensuring precise spatial alignment of sampling points with the object's true structure.

First, the initial sampling offsets $\{(\Delta x_i', \Delta y_i')\}$ are generated from the query feature $\mathbf{q}$ using a linear layer:
\begin{equation}
    \{(\Delta x_i', \Delta y_i') \mid i \in 0,1,...,N_{\text{p}}-1 \} = {\rm Linear}(\mathbf{q}),
\end{equation}
where $N_{\text{p}}$ denotes the number of sampling points.

Next, the final sampling offsets $\{(\Delta x_i, \Delta y_i)\}$ and sampling points $\{(x_i, y_i)\}$ are determined via a geometric transformation, which relocates sampling points to object corners and aligns them with the object's heading, as follows:
\begin{equation}
    \left[\!\!\begin{array}{c}
    \Delta x_i  \\ [0.5em]
    \Delta y_i  \\
    \end{array}\!\right] \! = \! 
    \left[\!\!\begin{array}{cc}
    \cos\theta\!\!\! & -\!\sin\theta\!\! \\ [0.5em]
    \sin\theta\!\!\! & \cos\theta\!\!\!\!  \\
    \end{array}\!\!\right]\!\!
    \left[\!\!\begin{array}{c}
    {\rm I}_{j} \cdot \frac{l}{2} + \Delta x_i' \\ [0.5em]
    {\rm I}_{j}' \cdot\! \frac{w}{2}\! + \Delta y_i' \\
    \end{array}\!\!\right] \!
\end{equation}

\vspace{-5pt}
\begin{equation}
    \left[\!\!\begin{array}{cc}
    x_i & y_i \!
    \end{array}\!\right] \! = \! 
    \left[\!\!\begin{array}{cc}
    x_c  & y_c
    \end{array}\!\!\right] \! + \!
    \left[\!\begin{array}{cc}
    \!\Delta x_i & \Delta y_i \!\! \\
    \end{array}\!\right],
\end{equation}
where $(I_j, I'_j) \,\in \{(1, 1), (1, -1), (-1, 1), (-1, -1)\}$. $j$ denotes the index of each corner and is the remainder of $i$ divided by four, the number of corners. The coordinates $(x_c, y_c)$ denote the object's center. $l$, $w$, and $\theta$ denote the length, width, and yaw of the predicted bounding box, respectively, obtained from the regression head of the previous transformer layer. In the first decoder layer, we initialize $l$, $w$, $\theta$ to zero.

The features are sampled using bilinear interpolation at each sampling point on the multi-modal BEV feature map $\mathbf{BEV}_{\text{fuse}}$, as follows:
\begin{equation}
    g_i = {\rm B}(\mathbf{BEV}_{\text{fuse}}, (x_i, y_i)) \in \mathbb{R}^{C},
\end{equation}
where ${\rm B}(\cdot)$ denotes the bilinear interpolation, and $g_i$ represents the sampled features at $(x_i, y_i)$.

\paragraph{Position-aware Feature Mixing}
Given the sampled features from different corners, the key challenge is how to decode them while maintaining their spatial relationships. While AdaMixer~\cite{gao2022adamixer} is designed for feature decoding, it struggles with the geometric transformations introduced during corner-aware sampling, which complicate the association between sampled features and their transformed sampling locations (see \cref{table:query_update}~(e)). To address this, we propose position-aware feature mixing, which incorporates positional embeddings that encode the transformed sampling offsets, allowing for more structured feature aggregation.

First, the sampling offsets are embedded as position vectors using sinusoidal position encoding \cite{vaswani2017attention}, followed by a linear layer. Then, the position-aware sampled features $G \in \mathbb{R}^{N_{\text{p}} \times C}$ are obtained by element-wise addition of the sampled feature $g_i$ and the position vector $e_i$:
\begin{align}
    e_i &= {\rm Linear}(\Phi_{pos}((x_i, y_i))) \in \mathbb{R}^{C} \\
    G &= \{(g_i + e_i) \mid i \in {0,1,...,N_{\text{p}}-1}\},
\end{align}
where $\Phi_{pos}(\cdot)$ denotes sinusoidal position encoding.

Subsequently, adaptive channel mixing is applied to $G$ using the dynamic weights $W_{c}$ generated from the query feature $\mathbf{q}$ to obtain the channel-mixed feature $G_c$:
\begin{align}
    W_{c} &= {\rm Linear}(\mathbf{q}) \in \mathbb{R}^{C\times C} \\
    G_c &= {\rm ReLU}({\rm LN}(G \times W_{c})).
\end{align}

Next, adaptive spatial mixing is applied to the spatial dimensions of $G_c$ using dynamic weights $W_{s}$ to obtain the spatial-mixed feature $G_{cs}$:
\begin{align}
    W_{s} &= {\rm Linear}(\mathbf{q}) \in \mathbb{R}^{N_{\text{p}}\times N_{\text{p}}} \\
    G_{cs} &= {\rm ReLU}({\rm LN}(G_{c}^{T} \times W_{s})).
\end{align}

Finally, the query feature is formulated as follows:
\begin{equation}
    \mathbf{q'} = \mathbf{q} + {\rm Linear}({\rm Flatten}(G_{cs})),
\end{equation}
where $\mathbf{q'}$ represents the refined query feature. By integrating positional embeddings at the feature level, this formulation enhances spatial awareness in query representation, ensuring more robust geometric reasoning in object queries.

%% file: sec/4_experiments.tex
\section{Experiments}

\input{table/table_1}
\subsection{Implementation Details}
For the image backbone, we use ResNet-50~\cite{resnet} with a resolution of $704\times256$ or V2-99~\cite{Lee2019VoVNetV2} with $1600\times640$ with FPN \cite{feature_pyramid_network}. The LiDAR backbone is VoxelNet~\cite{zhou2018voxelnet} with an ROI of $[-54.0m, 54.0m]$ in $(X, Y)$ and $[-5.0m, 3.0m]$ in $Z$, with a voxel size of $(0.075m, 0.075m, 0.2m)$. In ASAP, we use four sampling points for each grid cell. The multi-modal BEV feature map size is $180\times180$. Following \cite{yin2021center}, object groups in \ref{method:query_init} are defined as: (1) car, (2) truck, construction vehicle, (3) bus, trailer,  (4) barrier, (5) motorcycle, bicycle, (6) pedestrian, traffic cone. Each group contains 150 queries, resulting in a total of 900 queries, and the corner-aware sampling generates 16 sampling points per query, both of which were empirically determined. The transformer decoder has six layers, and the feature dimension is set to 256.

Our model is trained on 8 RTX 3090 GPUs with a batch size of 16. The model is trained end-to-end for 10 epochs using CBGS~\cite{zhu2019cbgs}, whereas GT sample augmentation~\cite{yan2018second} is applied for the first 9 epochs. The query denoising strategy \cite{li2022dn_detr} is also adopted. Gaussian Focal loss \cite{wang2022gaussian_focal_loss}, Focal loss \cite{ross2017focal_loss} and L1 loss are used for heatmap prediction, classification and regression, respectively. The AdamW~\cite{adamw} optimizer is adopted with a learning rate of $1 \times 10^{-4}$ and a weight decay of $1 \times 10^{-2}$ with a cyclical learning rate policy \cite{smith2017cyclical_lr}. No model ensemble or test-time augmentation is applied during inference.

\begin{figure*}[t]
    \centering
    \includegraphics[width=\linewidth]{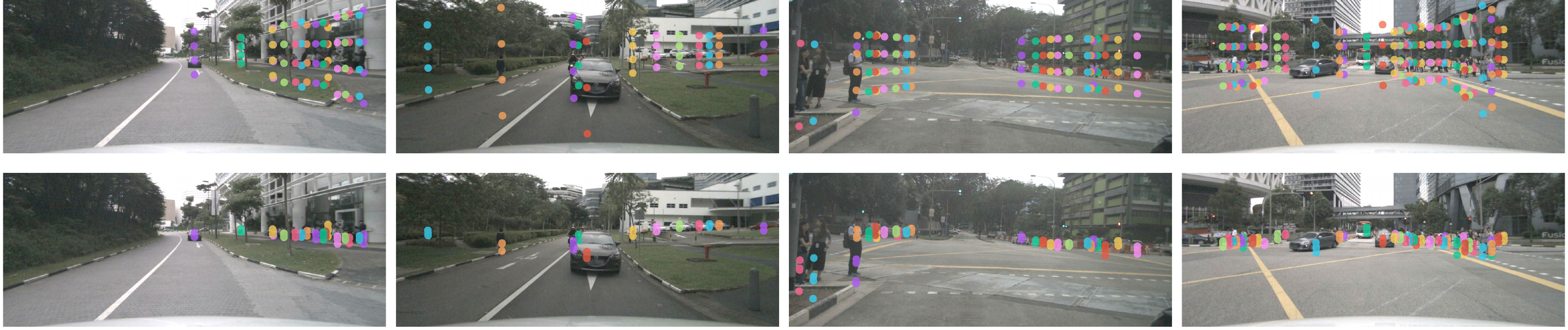}
    \caption{Visualization of projected sampling points for each object. The top row shows the projections of predefined 3D points, and the bottom row shows the projections of the points generated by the AS module. Points of each object are denoted by different colors.
    }
\label{fig:asap_proj}
\end{figure*}

\begin{figure}[t]
    \centering
    \begin{subfigure}{0.49\linewidth}
        \centering
        \includegraphics[width=\linewidth]{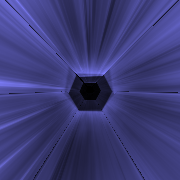}
        \caption{vanilla}
    \end{subfigure}
    \hfill
    \begin{subfigure}{0.49\linewidth}
        \centering
        \includegraphics[width=\linewidth]{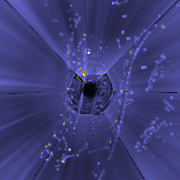}
        \caption{only AS}
    \end{subfigure}
    
    \vspace{0.1cm}
    
    \begin{subfigure}{0.49\linewidth}
        \centering
        \includegraphics[width=\linewidth]{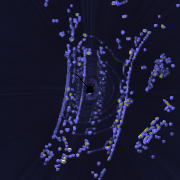}
        \caption{only AP}
    \end{subfigure}
    \hfill
    \begin{subfigure}{0.49\linewidth}
        \centering
        \includegraphics[width=\linewidth]{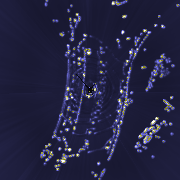}
        \caption{ASAP}
    \end{subfigure}
    
    \caption{Comparison of BEV feature maps from different methods: (a) Vanilla, (b) AS, (c) AP, and (d) ASAP. Vanilla shows less informative and unaligned features. AS improves feature representation using adaptive sampling. AP corrects ray-directional misalignment. ASAP integrates both AS and AP, leading to the most refined and well-aligned feature representation.
    }
    \label{fig:asap_bev}
\end{figure}

\subsection{Dataset and Metric}
Consistent with previous works~\cite{yin2024fusion, li2024gafusion, hu2023ea-lss, cai2023bevfusion4d, zhang2024sparselif, zhao2024simplebev}, we conduct extensive experiments on the nuScenes dataset~\cite{caesar2020nuscenes}, a large-scale benchmark for evaluating 3D object detection in autonomous driving. It consists of 1,000 scenes, each lasting 20$s$, divided into training, validation, and testing sets (700, 150, and 150 scenes, respectively). This dataset contains multimodal sensor data, including point clouds from a 32-beam LiDAR at 20 frames per second (fps), images from six cameras with a resolution of 1600×900 pixels at 12 fps, and data from five radars, providing a 360-degree view, with annotations are provided every 0.5$s$, resulting in 1.4 million annotated objects across 10 traffic categories.

Performance on this dataset is assessed using metrics like mean Average Precision (mAP), calculated over distance thresholds of 0.5, 1, 2, and 4m across all classes, and the nuScenes Detection Score (NDS), which offers a holistic evaluation by combining mAP with measures of translation, scale, orientation, velocity, and attribute errors.

\subsection{Comparison with State-of-the-art Methods}
As shown in \cref{table:nuscenes_val_test}, we compare EVT and its LiDAR-only model, EVT-L, with existing methods on the nuScenes validation and test sets. The multi-modal EVT achieves 75.3\% NDS and 72.6\% mAP, surpassing all previous approaches on both the nuScenes validation and test sets. In particular, it surpasses recent methods, such as UniTR~\cite{wang2023unitr} by 0.8\% NDS and 1.7\% mAP, MSMDFusion~\cite{jiao2023msmdfusion} by 1.3\% NDS and 1.1\% mAP, and SparseFusion~\cite{xie2023sparsefusion} by 1.5\% NDS and 0.6\% mAP.
Furthermore, EVT shows a performance improvement of 3.2\% NDS and 4.9\% mAP compared with the LiDAR-only model EVT-L. In contrast, TransFusion~\cite{Bai2022TransFusionRL} shows only a 1.5\% NDS and 3.4\% mAP improvement over its LiDAR-only model, TransFusion-L. This indicates that EVT effectively utilizes camera data through the proposed view transformation method.

We also compare EVT-L with the LiDAR-only versions of other multi-modal 3D object detectors. On the nuScenes validation set, EVT-L surpasses TransFusion~\cite{Bai2022TransFusionRL} and CMT~\cite{Yan2023cmt} by 1.6\% and 3.1\% NDS, respectively. EVT-L also demonstrates competitive performance on the test set. These results validate the effectiveness of the proposed query initialization and cross-attention mechanisms in enhancing 3D object detection.

\begin{table}
   \centering
   \begin{tabular}{l|cccc|cc|c}
      \toprule
       & LiDAR & Camera & AS & AP & NDS & mAP & FPS \\
      \midrule
      (a) & \checkmark & & & & 71.7 & 66.4 & 12.1 \\
      (b) & \checkmark & \checkmark & & & 72.7 & 69.1 & 8.5 \\
      (c) & \checkmark & \checkmark & \checkmark & & 73.5 & 70.6 & 8.5 \\
      (d) & \checkmark & \checkmark & \checkmark & \checkmark & 74.1 & 71.1 & 8.3 \\
      \bottomrule
   \end{tabular}
   \caption{Ablation study of the proposed VT method. The proposed ASAP shows a significant performance improvement compared to the vanilla VT method (b), which projects predefined 3D points.
   }
   \label{table:ablation_asap}
\end{table}

\begin{table}
   \centering
   \begin{tabular}{l|c|c|c|c|c|c}
        \toprule
        \multirow{2}{*}{Method} & \multicolumn{6}{c}{\# Decoder Layers \& NDS (\%)} \\ 
        \cmidrule{2-7}
        & 1 & 2 & 3 & 4 & 5 & 6  \\
        \midrule
        (a) Learnable Init.  & 56.8  & 64.4  & 67.5  & 68.7  & 68.9  & 69.6  \\
        (b) Heatmap Init.  & \textbf{70.4}  & 70.5  & 70.5  & 70.3  & 70.5  & 70.7  \\
        (c) Mixed init.  & 69.2  & 70.3  & 70.5  & 70.5  & 70.7  & 71.1  \\
        \quad + Group-wise  & 69.8  & \textbf{71.1}  & \textbf{71.3}  & \textbf{71.2}  & \textbf{71.4}  & \textbf{71.7}  \\
        \bottomrule
    \end{tabular}
    \caption{
        Comparison of query initialization strategies on the nuScenes validation set.
        (a) Fully learnable initialization, (b) Fully heatmap-based initialization, 
        (c) Our mixed initialization strategy: the first row denotes instance-wise mixed query selection and the second row denotes group-wise mixed query selection.
    }
    \vspace{-5pt}
    \label{table:ablation_query_init}
\end{table}

\begin{table*}
   \centering
   \begin{tabular}{l|c|c|cc|cc|cc}
      \toprule
      & Attention & Reference & scale & rotate & feature mixing & position-aware & NDS & mAP \\
      \midrule
      (a) & Standard & Center & & & & & 69.6 & 65.3 \\
      \midrule
      (b) & Deformable & Center & & & & & 70.0& 64.8 \\
      (c) & & & \checkmark & \checkmark & & & 70.5 & 65.2 \\
      \midrule
      (d) & Ours & Corner & & \checkmark & & & 71.3 & 65.7 \\
      (e) & & & & \checkmark & \checkmark & & 71.2 & 65.6 \\
      (f) & & & & \checkmark & \checkmark & \checkmark & \textbf{71.7} & \textbf{66.4} \\
      \bottomrule
   \end{tabular}
   \vspace{-5pt}
   \caption{Comparison of attention methods within the transformer decoder. The proposed geometry-aware cross-attention, which includes corner-aware sampling and position-aware feature mixing, achieves significant performance improvements.
   }
   \vspace{-10pt}
   \label{table:query_update}
\end{table*}

\begin{table}
   \centering
   \begin{tabular}{l|cc}
        \toprule
        Method & NDS & mAP \\
        \midrule
        ResNet-50 baseline (24 epochs)    & 47.8 & 37.2 \\
        \rowcolor{Gray}
        + geometry-aware cross-attention & \textbf{49.1} & \textbf{37.8} \\
        \midrule
        ResNet-50 baseline (90 epochs)    & 53.5 & 42.7 \\
        \rowcolor{Gray}
        + geometry-aware cross-attention & \textbf{54.7} & \textbf{43.4} \\
        \bottomrule
   \end{tabular}
   \vspace{-5pt}
   \caption{Impact of geometry-aware cross-attention on StreamPETR~\cite{wang2023stream_petr}. All models are trained in our experiments.}
   \vspace{-12pt}
   \label{table:gaqa_streampetr}
\end{table}

\subsection{Ablation Studies}
\label{sec:ablation}
In this section, we describe the validation of each component of the proposed method on the nuScenes validation set. Unless otherwise specified, all experiments are conducted using the proposed LiDAR-only model EVT-L, trained for 10 epochs with CBGS~\cite{zhu2019cbgs} and the denoising strategy~\cite{li2022dn_detr}.

\paragraph{Adaptive Sampling and Adaptive Projection}
In \cref{table:ablation_asap}, the ablation results for ASAP are obtained by retraining the entire model. In these experiments, ResNet \cite{resnet} is used as the backbone network with a resolution of $704 \times 256$. (a) shows the performance of EVT-L, the LiDAR-only model. In (b), the vanilla view transformation (VT) method, without LiDAR guidance, employs a 3D-to-2D projection of predefined 3D points for feature sampling.

In (c), AS achieves improvements of 0.8\% NDS and 1.5\% mAP compared to the vanilla VT method. The sampling points of the vanilla VT and AS are visualized in \cref{fig:asap_proj}. The sampling points of AS are adaptively generated in highly object-relevant regions of the image.

In (d), the entire ASAP achieves improvements of 1.4\% NDS and 2.0\% mAP compared to (b), while maintaining high efficiency with only a 3ms latency increase. The BEV feature maps of each component are visualized in \cref{fig:asap_bev}. ASAP effectively transforms image features into BEV space and resolves ray-directional misalignment.

\paragraph{Group-wise Mixed Query Selection}
As shown in \cref{table:ablation_query_init}, we conduct an ablation study on query initialization strategies. (a) fully learnable initialization, where both features and positions are learnable \cite{liu2022petr, liu2023petrv2, liu2023sparsebev, Yan2023cmt, hu2023fusionformer}.
(b) fully heatmap-based initialization, where positions are obtained from high-score heatmap keypoints, and features are sampled at those locations \cite{zhou2022centerformer, Bai2022TransFusionRL, yin2024fusion, zhang2024sparselif, wang2024mv2dfusion}.
(c) our mixed initialization strategy, which combines heatmap-derived positions with either instance-wise or group-wise embeddings.

These results highlight the importance of query initialization. Notably, fully learnable initialization without any prior information consistently yields the lowest performance. While (b) achieves the best performance in a single-layer transformer decoder, (c) outperforms all other strategies in multi-layer settings, improving NDS by over 1\% at the last layer. Additionally, group-wise embeddings allow each group to learn more generalized feature representations, leading to meaningful improvements.

\vspace{-8pt}
\paragraph{Geometry-aware Cross-Attention}
We ablate the corner-aware sampling method, as shown in \cref{table:query_update}~(a)-(d). (a) employs multi-head attention \cite{vaswani2017attention}, and (b)-(c) use deformable attention \cite{Zhu2020DeformableDetr}. Specifically, in (c), the sampling offsets are scaled based on the bounding box size and rotated according to the heading. (d), the proposed corner-aware sampling method, samples features from bounding box corners aligned with the heading, achieving improvements of 0.8\% NDS and 0.5\% mAP compared to (c).

In (e) and (f), AdaMixer~\cite{gao2022adamixer} fails to improve performance, as it does not effectively preserve the structured sampling introduced by corner-aware sampling. In contrast, our position-aware feature mixing explicitly incorporates positional embeddings, leading to improvements of 0.4\% NDS and 0.7\% mAP compared to (d). As a result, our proposed geometry-aware cross-attention achieves overall improvements of 1.2\% NDS and 1.2\% mAP.

Additionally, as shown in \cref{table:gaqa_streampetr}, we validate the applicability of geometry-aware cross-attention by integrating it into a camera-only 3D detector. Modifying only the cross-attention layers in StreamPETR~\cite{wang2023stream_petr} improves performance without further adjustments.

\begin{table}
   \centering
   \begin{tabular}{l|cc}
      \toprule
      Initial Query Formulation & NDS & mAP \\
      \midrule
      Bbox from regression head & 71.4 & 66.3 \\
      Bbox from learnable parameters & 70.9 & 66.0\\
      \rowcolor{Gray}
      BEV Reference Point & \textbf{71.7} & \textbf{66.4} \\
      \bottomrule
   \end{tabular}
   \vspace{-5pt}
   \caption{Comparison of the initial query formulations for the first layer of the transformer decoder.}
   \label{table:init_query_bbox}
\end{table}

\vspace{-10pt}
\paragraph{Initial Query Formulation for Transformer}
As described in \cref{method:query_update}, the corner-aware sampling method is applied starting from the second transformer layer. We explore two different approaches to extend the corner-aware sampling to all transformer layers. The first approach involves adding regression heads during query initialization in \cref{method:query_init} to predict bounding boxes, whereas the second approach uses learnable bounding boxes. However, as shown in \cref{table:init_query_bbox}, neither approach resulted in any noticeable performance gains, and the experiment with learnable bounding boxes even led to performance degradation.

%% file: table/table_1.tex
\setlength{\tabcolsep}{4pt}
\begin{table*}[t]
   \centering
   \begin{tabular}{l|c|cc|cc}
      \toprule
      Method & Modality & NDS (\textit{val}) & mAP (\textit{val}) & NDS (\textit{test}) & mAP (\textit{test}) \\
      \midrule
      UVTR-L \cite{Li2022uvtr}                           & L & 67.7 & 60.9 & 69.7 & 63.9 \\
      TransFusion-L \cite{Bai2022TransFusionRL}          & L & 70.1 & 65.1 & 70.2 & 65.5 \\
      FocalFormer3D-L \cite{chen2023focalformer3d}       & L & - & - & \textbf{72.6} & \textbf{68.7} \\
      CMT-L \cite{Yan2023cmt}                            & L & 68.6 & 62.4 & 70.1 & 65.3 \\
      \rowcolor{Gray}
      \textbf{EVT-L} (Ours)                              & L & \textbf{71.7} & \textbf{66.4} & 72.1 & 67.7 \\
      \midrule
      MVP \cite{yin2021multimodal_mvp}                   & LC & 70.8 & 67.1 & 70.5 & 66.4 \\
      UVTR \cite{Li2022uvtr}                             & LC & 70.2 & 65.4 & 71.1 & 67.1 \\
      AutoAlignV2 \cite{chen2022autoalignv2}             & LC & 71.2 & 67.1 & 72.4 & 68.4 \\
      TransFusion \cite{Bai2022TransFusionRL}            & LC & 71.3 & 67.5 & 71.7 & 68.9 \\
      DeepInteraction \cite{yang2022deepinteraction}     & LC & 72.6 & 69.9 & 73.4 & 70.8 \\
      BEVFusion \cite{liu2023bevfusion}                  & LC & 71.4 & 68.5 & 72.9 & 70.2 \\
      Objectfusion \cite{cai2023objectfusion}            & LC & 72.3 & 69.8 & 73.3 & 71.0 \\
      FocalFormer3D \cite{chen2023focalformer3d}         & LC & 71.1 & 66.5 & 73.9 & 71.6 \\
      CMT \cite{Yan2023cmt}                              & LC & 72.9 & 70.3 & 74.1 & 72.0 \\
      BEVFusion4D-S \cite{cai2023bevfusion4d}            & LC & 72.9 & 70.9 & 73.7 & 71.9 \\
      SparseFusion \cite{xie2023sparsefusion}            & LC & 72.8 & 70.4 & 73.8 & 72.0 \\
      MSMDFusion \cite{jiao2023msmdfusion}               & LC & - & - & 74.0 & 71.5 \\
      UniTR \cite{wang2023unitr}                         & LC & 73.3 & 70.5 & 74.5 & 70.9 \\
      FusionFormer \cite{hu2023fusionformer}             & LCT & 74.1 & 71.4 & 75.1 & 72.6 \\
      \rowcolor{Gray}
      \textbf{EVT} (Ours)                                & LC & \textbf{74.6} & \textbf{72.1} & \textbf{75.3} & \textbf{72.6} \\
      \bottomrule
   \end{tabular}
   \vspace{-5pt}
   \caption{Performance comparison on the nuScenes validation and test sets. The results are obtained without model ensemble or test-time augmentation. ‘L’, ‘C’ and ‘T’ denote LiDAR, camera and temporal fusion, respectively.}
   \vspace{-10pt}
   \label{table:nuscenes_val_test}
\end{table*}

%% file: sec/5_conclusions.tex
\section{Conclusion}
We propose EVT, a novel multi-modal 3D object detector based on BEV representation, enhancing both efficiency and accuracy. Our method introduces ASAP, an efficient LiDAR-camera fusion method that leverages LiDAR guidance for accurate view transformation. Additionally, the proposed group-wise mixed query selection improves initial feature representation through shared embeddings. The geometry-aware cross-attention refines queries using geometric properties and can be easily extended to other models. We expect EVT to provide valuable insights into multi-modal 3D object detection.